\documentclass[runningheads]{llncs}
\usepackage[T1]{fontenc}
\usepackage{graphicx,verbatim}
\usepackage{booktabs}
\usepackage{multirow}
\usepackage{tablefootnote}
\usepackage{amssymb}
\usepackage[table]{xcolor}
\usepackage{pifont}

\usepackage{cite}
\usepackage{orcidlink}

\makeatletter
\renewcommand*{\@fnsymbol}[1]{\textbf{\normalsize *}}
\newcommand{\printfnsymbol}[1]{%
  \textsuperscript{\@fnsymbol{#1}}%
}
\makeatother

\begin{document}
\title{Exploiting Longitudinal Context in Clinician-Verified Interactive Lesion Tracking}
\titlerunning{Longitudinal Context in Verified Lesion Tracking}
%
\author{Yannick Kirchhoff\inst{1,2,3}\thanks{Contributed equally. Each co-first author may list themselves as lead on their CV.} \and
Maximilian Rokuss\inst{1,2,3}\printfnsymbol{1} \and
Daniel Philipp Mertens\inst{5} \and
David Füller\inst{5} \and
Benjamin Hamm\inst{1,4} \and
Andreas Schreyer\inst{5} \and
Oliver Ritter\inst{5} \and
Klaus Maier-Hein\inst{1,4,6}
}
\authorrunning{Y. Kirchhoff, M. Rokuss et al.}
\institute{German Cancer Research Center (DKFZ) Heidelberg, Division of Medical Image Computing, Germany
\and
Faculty of Mathematics and Computer Science, Heidelberg University, Germany
\and
HIDSS4Health -- Helmholtz Information and Data Science School for Health, Karlsruhe/Heidelberg, Germany
\and
Medical Faculty, Heidelberg University, Germany
\and
University Hospital Brandenburg an der Havel, Brandenburg Medical School Theodor Fontane, Germany
\and
Pattern Analysis and Learning Group, Department of Radiation Oncology, Heidelberg University Hospital, Germany\\
\email{\{yannick.kirchhoff,maximilian.rokuss\}@dkfz-heidelberg.de}
}
  
\maketitle              

\begin{abstract}
Tracking tumor lesions across serial CT scans is essential for oncological response assessment. Existing automated methods face a fundamental trade-off: end-to-end trackers achieve high automation but offer no opportunity to correct silent tracking failures, while decoupled registration–segmentation pipelines permit user verification yet discard the lesion's prior appearance, limiting accuracy in ambiguous cases. In this work, we propose a \textit{Verified Tracking} paradigm: a clinician verifies a registration-proposed prompt, which the model leverages alongside the baseline lesion appearance to resolve segmentation ambiguities. We present a unified framework combining early spatial prompt fusion with latent temporal difference weighting for longitudinally-informed segmentation. To address data scarcity, we leverage large-scale synthetic pretraining, proving essential for exploiting longitudinal context, improving performance by up to 4.5 Dice points over training from scratch. Our approach secured first place in the MICCAI autoPET IV challenge. We further curate and release \textit{PanTrack,} a new longitudinal pancreatic cancer benchmark, to assess out-of-distribution generalization. Experiments show that our model outperforms prior work in both fully automatic and the proposed verified tracking setting offering a clinically safe middle ground between automation and control. Code, model and dataset will be released at \url{https://github.com/MIC-DKFZ/LongiSeg}.

\end{abstract}

\section{Introduction}

Longitudinal imaging is the cornerstone of oncological response assessment. With cancer incidence projected to rise 47\% by 2040~\cite{bray2024global} and CT examination volumes increasing steadily~\cite{rocholl2025unstable}, radiologists face growing pressure to evaluate serial scans efficiently. Treatment response evaluation, typically governed by RECIST 1.1 guidelines~\cite{recist}, requires solving two distinct subtasks per lesion: \textit{retrieval}, identifying the same structure in a follow-up scan, and \textit{delineation}, measuring its volume change. Both remain predominantly manual, causing substantial reading time and inter-observer variability~\cite{hering2024improving,jacobs2021assisted,lu2021randomized}.

\noindent Existing automated approaches address this problem in complementary yet incomplete ways. \textbf{Point-only trackers}~\cite{yan2022sam,vizitiu2023multi} retrieve lesion centers, however, without volumetric delineation. \textbf{End-to-end trackers}~\cite{Rokuss_2025_CVPR} automate both retrieval and segmentation using a single baseline click, but operate as "black boxes": if the model tracks an incorrect structure or misses a splitting lesion, the resulting segmentation is clinically invalid with no opportunity for correction. \textbf{Decoupled registration-segmentaion pipelines}~\cite{hering2021whole} would permit verification of the propagated point but suffer twofold: (1) registration errors frequently displace prompts beyond the tolerance of segmentation models trained on well-centered clicks~\cite{rocholl2025unstable,scribbleprompt,nnInteractive,du2024segvol}, and (2) treating follow-up scans in isolation discards the \textit{longitudinal prior} (baseline appearance) needed to resolve ambiguous findings. Finally, \textbf{automated longitudinal models}~\cite{longiseg, szeskin2023liver, wu2023coactseg} leverage this prior lesion appearance for ehnanced segmentation but lack promptability, preventing user interaction or correction at all.\\

\noindent To bridge these methodological gaps, we argue that clinical deployment requires satisfying three criteria simultaneously: (i) explicit use of the baseline lesion as a longitudinal prior, (ii) a clinician-correctable mechanism to guarantee correspondence when tracking fails, and (iii) robustness to off-center prompts caused by registration or rapid correction. Because existing methods satisfy at most two, we propose a novel \textit{Verified Tracking} paradigm: registration proposes a follow-up location, a clinician verifies/corrects it, and the model segments using both the verified prompt and baseline context. This eliminates retrieval failures through minimal oversight, freeing the model to focus entirely on longitudinally-informed delineation. We present a unified framework for this workflow, combining early pixel-level prompt fusion~\cite{nnInteractive} with latent-space temporal difference weighting~\cite{longiseg}. Critically, we identify \textit{synthetic longitudinal pretraining} as a decisive enabler: without sufficient multi-timepoint data, longitudinal architectures collapse to single-timepoint shortcuts, ignoring the baseline scan entirely. Finally, to address longitudinal data scarcity, we curate and publicly release \textit{PanTrack} as a dedicated out-of-distribution benchmark. As public datasets with consistent, lesion-level instance annotations across multiple timepoints remain largely unavailable, this release fills a major gap, providing a valuable new resource for lesion tracking model development. Our key contributions are:

\begin{enumerate}
    \item \textbf{Verified Tracking formulation}: We formalize a workflow where follow-up prompts are registration-proposed and optionally corrected, offering a clinically safe middle ground between automation and control.
    \item \textbf{Longitudinal promptable segmentation model}: A unified architecture combining early prompt fusion and latent temporal difference weighting. Enhanced by promptable large-scale synthetic pretraining, our method won the MICCAI autoPET IV challenge, outperforming the state-of-the-art in automatic and verified tracking.
    \item \textbf{The \textit{PanTrack} benchmark}: To address multi-timepoint data scarcity, we publicly release 161 curated longitudinal CT scans (45 pancreatic cancer patients) to provide a rigorous out-of-distribution testbed for cross-domain tracking generalization and a novel model development resource.
\end{enumerate}

 \section{Method}

\begin{figure}[t]
\centering
    \includegraphics[width=1\textwidth]{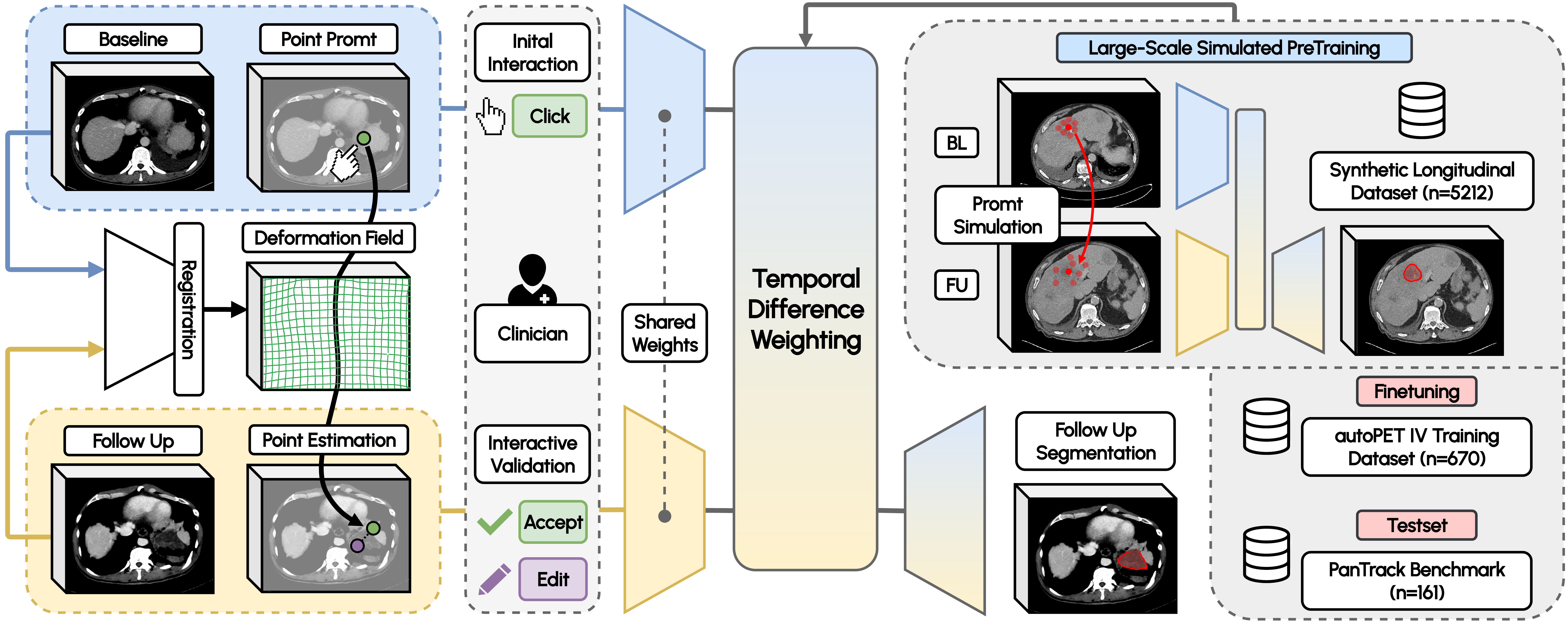}
\caption{\textbf{Overview of our framework.} The registration proposes a candidate follow-up prompt which the clinician verifies or corrects. A shared-weight encoder processes both (image, prompt) pairs and in latent space a Difference Weighting Block fuses their features by explicitly attending to temporal change before the decoder produces the longitudinally-informed follow-up segmentation. Prior, the model is pretrained on a large-scale synthetic longitudinal corpus with simulated prompts for both timepoints.}
\label{fig:overview}
\end{figure}

\subsection{Problem Formulation}

Given a baseline CT scan $I_0$ with a known lesion center $p_0$ and a follow-up CT scan $I_t$, our goal is to produce a volumetric segmentation of the corresponding lesion in $I_t$, conditioned on a clinician-verified follow-up point prompt $p_t$.

\subsection{Verified Tracking via Registration}

We decouple the longitudinal tracking workflow into two stages: (1)~a \textit{retrieval} step, handled jointly by a registration model and the clinician, and (2)~a \textit{delineation} step, performed by our segmentation network. For the retrieval step, we apply uniGradICON~\cite{uniGradICON}, a registration foundation model trained across diverse anatomical regions, to estimate a deformation field $\phi: I_0 \to I_t$, yielding a candidate follow-up prompt:
\begin{equation}
    \hat{p}_t = \phi(p_0).
\end{equation}
The clinician views $\hat{p}_t$ superimposed on $I_t$ and either accepts it or provides a corrected $p_t$. This verification step eliminates retrieval failures while preserving a high degree of clinical automation. For the autoPET IV dataset, the provided registration-propagated center points~\cite{kuestner2025longitudinalct} serve directly as $\hat{p}_t$.

\subsection{Segmentation Network Architecture}

Unlike decoupled pipelines that segment the follow-up in isolation, our model conditions delineation on both $(I_t, p_t)$ \emph{and} $(I_0, p_0)$, explicitly learning to use baseline appearance to resolve ambiguities that a single-timepoint model cannot. Our segmentation network must simultaneously handle two distinct integration challenges: incorporating a \textit{spatial point prompt} and leveraging \textit{baseline appearance}. These two signals empirically demand fusion at different levels of abstraction: spatial prompts are most effective at the image input level~\cite{nnInteractive,scribbleprompt}; longitudinal context requires latent-space integration with an explicit inductive bias towards temporal differences to prevent the network from ignoring the baseline branch altogether~\cite{longiseg,denner2020spatio}. We combine both in a Residual Encoder U-Net~\cite{nnunet_revisited} with the hybrid fusion design below.\\

\noindent\textbf{Early Prompt Fusion.}
After verified registration, we extract Volumes of Interest (VOIs) centered at $p_0$ and $p_t$ from $I_0$ and $I_t$, respectively. Following prior findings that prompt encoding is most effective at the input level~\cite{nnInteractive, scribbleprompt}, we concatenate image and prompt channel-wise:
\begin{equation}
    X_0 = [I_0,\; G(p_0)], \quad X_t = [I_t,\; G(p_t)],
\end{equation}
where $G(p)$ denotes a Gaussian heatmap with $\sigma=1$ centered at $p$, rescaled to unit value at the center. Crucially, $p_t$ does not need to lie precisely at the lesion center; residual registration error or clinical corrections are explicitly anticipated. Both pairs are then processed by a \textit{shared-weight} encoder, yielding multi-scale features $\{x_0^l\}$ and $\{x_t^l\}$ at each resolution level $l$.\\

\noindent\textbf{Latent Temporal Fusion via Difference Weighting.}
Naive channel-wise concatenation of multi-timepoint features risks the network primarily attending to a single timepoint, effectively collapsing to cross-sectional behavior~\cite{longiseg,denner2020spatio}. To impose an explicit longitudinal inductive bias, we apply a Difference Weighting Block (DWB)~\cite{longiseg} at all U-Net skip connections. For baseline and follow-up features ($x_0^l$, $x_t^l$), the DWB computes:
\begin{equation}
    {x'}_t^{\,l} = x_t^l \;\times\; \mathrm{InstNorm}\!\left(x_t^l - x_0^l\right) + x_t^l.
\end{equation}
The normalized feature difference acts as an attention map, gating $x_t^l$ to emphasize regions of longitudinal change. This lightweight operation runs at all resolutions without architectural overhead. The temporally-informed features $\{{x'}_t^l\}$ are passed to the decoder to generate the follow-up segmentation. To our knowledge, this is the first framework to combine early prompt fusion with latent temporal fusion, applying each mechanism precisely where it is most effective.\\

\noindent\noindent\textbf{Synthetic Longitudinal Pretraining.}
Real annotated multi-timepoint data is scarce, and without sufficient
training data even a longitudinal architecture can collapse
to cross-sectional behavior, ignoring the baseline scan entirely.
We address this by extending the synthetic corpus of~\cite{Rokuss_2025_CVPR}
to the promptable setting: 2{,}606 CT pairs with synthetic follow-ups
generated via anatomy-informed deformation fields~\cite{kovacs2023anatomy} simulating tumor
growth, shrinkage, and acquisition variability are paired with full
prompt simulation for both timepoints.\\

\noindent\textbf{Prompt Simulation.} To instill robustness against imprecise user interactions and registration errors, training prompts are generated via a 50/50 split: half are sampled from the ground-truth mask with probabilities weighted by $1/d^2$, where $d$ is the distance to the centroid, and half are derived from the registered follow-up point, which may even fall outside the lesion.

\section{Datasets}


\noindent \textbf{autoPET/CT IV Dataset.} This dataset comprises longitudinal whole-body CT scans from 285 melanoma patients undergoing therapy response assessment~\cite{kuestner2025longitudinalct} totalling 670 images. Each patient has at least one baseline and follow-up scan acquired during portal-venous phase on multiple Siemens scanners with standardized protocols. Two radiologists manually segmented all tumor lesions across timepoints with side-by-side verification to establish correspondence. The dataset includes challenging scenarios (splitting, merging, vanishing lesions) and provides pre-computed lesion centers at both timepoints, with registration-based propagated clicks simulating the \textit{Verified Tracking} workflow.\\

\noindent \noindent \textbf{The PanTrack Dataset.} To evaluate generalization to a different anatomical domain, we curated \textit{PanTrack}, comprising 45 patients with pancreatic adenocarcinoma amounting to 161 CT examinations (2–11 per patient; mean 3.6). All scans were acquired on identical Siemens protocols (portal-venous phase) at a single institution. An experienced radiologist with pancreatic imaging expertise manually segmented all pancreatic lesions across timepoints, including hepatic metastases if present. The cohort represents diverse trajectories: some patients underwent long-term stable chemotherapy, others showed rapid progression. Unlike melanoma metastases, pancreatic lesions exhibit fuzzy boundaries and subtle soft-tissue contrast, providing complementary evaluation under different radiological characteristics. This dataset is publicly released upon acceptance.
\section{Experiments and Results}

\subsection{Experimental Setup}

We develop and train our model exclusively on the autoPET IV dataset~\cite{kuestner2025longitudinalct}, optimizing a combined Dice and cross-entropy loss via SGD for 1000 epochs. We randomly reserve one-third of the patients as a held-out test set, splitting the remaining cohort into an 80/20 training and validation split. All architectural decisions and hyperparameters were fixed prior to evaluation on the held-out test set and PanTrack. Crucially, PanTrack was completely excluded from any form of training or model selection, serving as a rigorous, entirely unseen out-of-distribution (OOD) benchmark. 

\noindent\textbf{Evaluation Protocol.} We evaluate under two tracking paradigms reflecting distinct clinical workflows. In the \textbf{Automatic Tracking} setting, only the baseline prompt $p_0$ is provided; the follow-up prompt $\hat{p}_t$ is generated fully automatically via uniGradICON registration. In the \textbf{Verified Tracking} setting, the ground-truth follow-up lesion centroid is provided, simulating a workflow where a clinician has accepted or swiftly corrected the registration-proposed location. All models receive prompts appropriate to the respective paradigm. Importantly, verified tracking eliminates catastrophic retrieval failures by design; remaining errors are purely delineation errors, which we quantify via Dice Similarity Coefficient (DSC) and Normalized Surface Distance (NSD). We additionally report the lesion detection rate (LDR) to assess detection validity.

\noindent\textbf{Baselines.} For automatic tracking, we compare against a registration-based decoupled pipeline (Hering et al.~\cite{hering2021whole}, reimplemented with uniGradICON), an end-to-end tracker (LesionLocator~\cite{Rokuss_2025_CVPR}), and nnInteractive~\cite{nnInteractive} prompted with the registration-propagated center. For verified tracking, we compare against interactive foundation models: nnInteractive~\cite{nnInteractive}, SegVol~\cite{du2024segvol}, and the official Universal Lesion Segmentation (ULS) model~\cite{uls_challenge}. While off-the-shelf foundation models are evaluated zero-shot on autoPET, potentially giving our trained model an in-domain advantage, the PanTrack dataset provides a strictly fair, zero-shot evaluation ground for all methods.

\subsection{Model Development: Unlocking the Longitudinal Prior}

We perform a systematic ablation study on the autoPET IV validation set to isolate the contributions of our architectural design and training strategy (Tab.~\ref{tab:results}).\\ 

\begin{table}[t]
\centering
\caption{\textbf{Ablation study.} Naive longitudinal concatenation from scratch (row 4) actually underperforms the single-timepoint baseline (row 1). While synthetic pretraining activates the architecture, Difference Weighting (DW) is essential to fully exploit the temporal prior. Best results in \textbf{bold}, second-best \underline{underlined}.}

\label{tab:results}
\setlength{\tabcolsep}{4pt} 
\begin{tabular}{l ccc ccc} 
\toprule
\textbf{Architecture} & \shortstack{\textbf{Temporal}\\\textbf{Fusion}} & \shortstack{\textbf{Prompt}\\\textbf{Sim.}} & \shortstack{\textbf{Pre-}\\\textbf{train}} & \textbf{DSC}~$\uparrow$ & \textbf{NSD}~$\uparrow$ & \textbf{LDR}~$\uparrow$ \\
\midrule
\multirow{2}{*}{\shortstack{Single\\Timepoint}} & -- & \checkmark & -- & 55.8 & 69.6 & 76.1 \\
 & -- & \checkmark & \checkmark & \underline{56.9} & 69.1 & 75.8 \\
\midrule
 & Concat. & -- & -- & 45.8 & 59.0 & 66.2 \\
 & Concat. & \checkmark & -- & 54.0 & 67.0 & 76.0 \\
 & Concat. & \checkmark & \checkmark & 56.5 & \underline{70.6} & \underline{77.7} \\
\rowcolor{gray!10}
\multirow{-4}{*}{\cellcolor{white}Longitudinal} & \textbf{Diff. Weighting} & \checkmark & \checkmark & \textbf{58.5} & \textbf{72.3} & \textbf{78.8} \\
\bottomrule
\end{tabular}
\end{table}

\noindent\textbf{The Failure of Naive Longitudinal Fusion.} Theoretically, providing the baseline scan should give the network strictly more information to resolve ambiguous boundaries. However, when trained from scratch, the naive longitudinal early fusion model (54.0 DSC) actually underperforms the single-timepoint baseline (55.8 DSC), confirming simply concatenating the inputs is insufficient.\\
\noindent\textbf{Pretraining Activates Longitudinal Functionality.} Introducing large-scale synthetic longitudinal pretraining~\cite{Rokuss_2025_CVPR} fundamentally changes the network's behavior. Pretraining largely prevents the cross-sectional collapse, allowing the naive longitudinal model to finally surpass the single-timepoint baseline in boundary accuracy (70.6 vs. 69.1 NSD) and detection rate (77.7 vs. 75.8 LDR), proving that synthetic priors are essential for learning temporal correspondences.\\
\noindent\textbf{Difference Weighting Maximizes the Temporal Prior.} While pretraining activates the architecture, naive channel-wise concatenation remains a suboptimal fusion strategy. Replacing it with Difference Weighting (DW) explicitly forces the model to attend to longitudinal changes by computing normalized feature differences in latent space. This combined approach, i.e. synthetic pretraining to prevent collapse, and DW to provide the correct structural inductive bias, yields a decisive performance leap (58.5 DSC, 72.3 NSD), fully unlocking the value of longitudinal context.\\
\noindent\textbf{Robustness via Prompt Simulation.} Finally, we note that training strictly on perfect center-point prompts causes catastrophic degradation on realistic, slightly off-center registration prompts (45.8 DSC). Dynamically sampling simulated prompts during training is a critical requirement to ensure the localization robustness demanded by the \textit{Verified Tracking} paradigm.



\subsection{Comparison with State-of-the-Art}

We evaluate our final model on both the autoPET IV test set and the OOD PanTrack dataset. Table~\ref{tab:main_results} details the results against established baselines.\\

\begin{table}[t]
\caption{\textbf{Comparison on held-out test sets.} Methods are grouped by tracking paradigm: automatic (prompted via registration) and verified (prompted via centroid). \textbf{Bold} indicates the best result per dataset with mean\,{\scriptsize ±std} obtained via bootstrapping.}
\label{tab:main_results}
\centering
\setlength{\tabcolsep}{3pt}
\begin{tabular}{cl ccc ccc}
\toprule
 & & \multicolumn{3}{c}{\textbf{autoPET IV (test)}} & \multicolumn{3}{c}{\textbf{PanTrack (ours)}} \\
\cmidrule(lr){3-5}\cmidrule(lr){6-8}
 & \textbf{Method} & DSC~$\uparrow$ & NSD~$\uparrow$ & LDR~$\uparrow$
                  & DSC~$\uparrow$ & NSD~$\uparrow$ & LDR~$\uparrow$ \\
\midrule
 & Hering et al.~\cite{hering2021whole}
   & 56.6{\scriptsize ±2.4} & 65.9{\scriptsize ±2.7} & 76.7{\scriptsize ±2.9}
   & 49.9{\scriptsize ±3.2} & 41.3{\scriptsize ±2.9} & 83.1{\scriptsize ±4.7} \\
 & nnInteractive~\cite{nnInteractive}
   & 43.3{\scriptsize ±2.7} & 49.5{\scriptsize ±3.0} & 61.3{\scriptsize ±3.4}
   & 34.3{\scriptsize ±3.5} & 27.2{\scriptsize ±2.7} & 59.3{\scriptsize ±5.7} \\
 & LesionLocator~\cite{Rokuss_2025_CVPR}
   & 44.1{\scriptsize ±2.9} & 51.7{\scriptsize ±3.2} & 61.1{\scriptsize ±3.7}
   & 51.7{\scriptsize ±3.4} & 41.3{\scriptsize ±3.3} & 84.0{\scriptsize ±4.2} \\
\rowcolor{gray!10}
\multirow{-4}{*}{\cellcolor{white}\rotatebox[origin=c]{90}{\shortstack{Automatic}}}
 & \textit{\textbf{Ours}}
   & \textbf{60.7}{\scriptsize ±2.6} & \textbf{69.5}{\scriptsize ±2.9} & \textbf{77.7}{\scriptsize ±3.2}
   & \textbf{58.2}{\scriptsize ±3.0} & \textbf{48.6}{\scriptsize ±3.0} & \textbf{93.7}{\scriptsize ±3.0}\\
\midrule
 & SegVol~\cite{du2024segvol}
   & 45.8{\scriptsize ±2.0} & 57.5{\scriptsize ±2.0} & 79.4{\scriptsize ±2.8}
   & 43.7{\scriptsize ±2.3} & 33.4{\scriptsize ±2.6} & 87.4{\scriptsize ±4.2} \\
 & ULS Model~\cite{uls_challenge}
   & 68.7{\scriptsize ±1.6} & 79.7{\scriptsize ±1.9} & 93.3{\scriptsize ±1.7}
   & 53.4{\scriptsize ±2.7} & 42.8{\scriptsize ±2.9} & \textbf{95.5}{\scriptsize ±2.4} \\
 & nnInteractive~\cite{nnInteractive}
   & 59.6{\scriptsize ±2.4} & 66.6{\scriptsize ±2.7} & 80.6{\scriptsize ±2.8}
   & 52.8{\scriptsize ±3.1} & 42.2{\scriptsize ±2.9} & 93.5{\scriptsize ±2.9} \\
\rowcolor{gray!10}
\multirow{-4}{*}{\cellcolor{white}\rotatebox[origin=c]{90}{\shortstack{Verified}}}
 & \textit{\textbf{Ours}}
   & \textbf{73.7}{\scriptsize ±1.5} & \textbf{84.0}{\scriptsize ±1.7} & \textbf{95.5}{\scriptsize ±1.2}
   & \textbf{60.0}{\scriptsize ±2.8} & \textbf{49.7}{\scriptsize ±2.9} & 94.7{\scriptsize ±2.7} \\
\bottomrule
\end{tabular}
\end{table}


\noindent\textbf{Automatic Tracking.} While designed for human-in-the-loop verification, our method establishes state-of-the-art performance even fully automatically. The sharp decline of nnInteractive (43.3 DSC) highlights its vulnerability to off-center prompts. Conversely, our model is highly robust to residual registration errors, outperforming both decoupled pipelines (Hering et al.) and end-to-end trackers (LesionLocator) without requiring human intervention.

\begin{figure}[t]
\centering
    \includegraphics[width=1\textwidth]{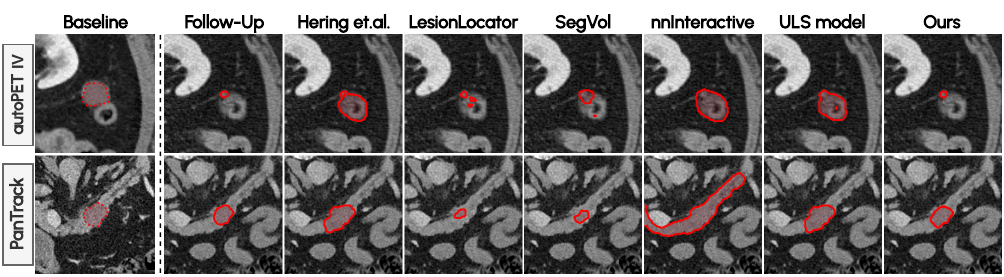}
\vspace{-0.7cm}
\caption{\textbf{Qualitative comparison} on autoPET IV (top) and PanTrack (bottom). Single-timepoint baselines struggle with ambiguous lesion volumes. In the top row, a shrinking lesion borders the colon; without the baseline appearance as context, competing models fail to isolate the correct structure when prompted near the organ boundary.}
\label{fig:qualitative}
\end{figure}

\noindent\textbf{Verified Tracking.} Clinician verification improves all methods, confirming that localization is the primary tracking bottleneck. Given identical verified prompts, our model achieves peak performance on both test sets (73.7 autoPET / 60.0 PanTrack DSC), substantially outperforming prior promptable models. This margin isolates the value of the longitudinal prior: our model can leverage the baseline appearance to delineate ambiguous boundaries that single-timepoint models fail to resolve (see Fig.~\ref{fig:qualitative}).

\noindent\textbf{OOD Generalization on PanTrack.} Despite PanTrack's different scanners, institutions, and fuzzy lesion boundaries, our method maintains its superiority. Strikingly, our model's \textit{automatic} tracking performance (58.2 DSC) not only closely approaches our \textit{verified} performance (60.0 DSC), but outright surpasses the \textit{verified} results of all competing models (e.g., ULS at 53.4 DSC). This proves our framework delivers highly reliable automated tracking "in the wild" while natively preserving the safety of optional clinician correction.

\section{Conclusion and Outlook}

We present a robust framework for longitudinally-informed lesion tracking that secured first place in the MICCAI autoPET IV challenge and demonstrates state-of-the-art generalization on \textit{PanTrack}, a novel out-of-distribution dataset. We propose ``Verified Tracking'' as a clinically viable middle ground for standard RECIST 1.1 workflows. By requiring clinicians to merely verify or correct a single follow-up point, this paradigm averts catastrophic retrieval failures and gracefully handles complex topological changes like splitting or vanishing lesions. Once anchored, our architecture leverages explicit temporal fusion and synthetic pretraining to fully exploit the baseline appearance, significantly improving delineation accuracy. While our framework currently assumes a known baseline lesion, a step readily automated by off-the-shelf detectors, it successfully isolates and solves the critical bottleneck of longitudinal correspondence. Building on our strong zero-shot OOD performance, future work will focus on prospective clinical reader studies and exploring richer prompt modalities beyond points, such as free-text descriptions of lesion characteristics~\cite{Rokuss_2026_CVPR}. To foster further research in generalizable tracking, we publicly release the \textit{PanTrack} dataset, along with our code and model weights.

\begin{credits}
\subsubsection{\ackname} This work was partly funded by the Helmholtz Information and Datascience School (HIDSS) and Helmholtz Imaging (HI), platforms of the Helmholtz Incubator on Information and Data Science. Supported by the Helmholtz Foundation Model Initiative (HFMI) through the pilot project THRP (The Human Radiome Project). Funded by the Deutsche Forschungsgemeinschaft (DFG, German Research Foundation) – 402688427. This project was funded within the DKTK Heidelberg Seed Funding 25 program. M.R. is funded through a Google PhD Fellowship.

\subsubsection{\discintname} The authors have no competing interests to declare. 
\end{credits}

%
%
%
\bibliographystyle{splncs04}
\bibliography{main}
\end{document}